\documentclass[11pt, a4paper, logo, copyright]{eai}
\usepackage[]{mdframed}
\usepackage{dblfloatfix}

\usepackage{amsmath,amsfonts,bm}
\usepackage{multirow}
\usepackage{subcaption}
\usepackage{wrapfig}

\def\eqref#1{equation~\ref{#1}}

\def\1{\bm{1}}

\DeclareMathAlphabet{\mathsfit}{\encodingdefault}{\sfdefault}{m}{sl}
\SetMathAlphabet{\mathsfit}{bold}{\encodingdefault}{\sfdefault}{bx}{n}

\usepackage{listings}
\usepackage{hyperref}
\usepackage{url}
\usepackage{graphicx}
\usepackage{amsmath} %
\usepackage{mathrsfs} %
\usepackage{etoolbox}
\usepackage{cleveref}
\usepackage{tcolorbox}
\usepackage{colortbl}
\usepackage{booktabs}       %
\usepackage{amsfonts}       %
\usepackage{nicefrac}       %
\usepackage{microtype}      %
\usepackage{caption}
\usepackage{xspace} % 
\usepackage{subcaption}
\usepackage{algorithm}
\usepackage{algorithmic}
\usepackage{multirow}
\usepackage{subcaption}
\usepackage{lipsum}

\usepackage{adjustbox}

\usepackage{booktabs} %
\usepackage{enumitem}%
\setlist[itemize]{noitemsep, topsep=0pt}
\usepackage{enumitem,kantlipsum}

\usepackage{multicol,multirow} 
\usepackage{graphicx}    
\usepackage{pifont}
\usepackage{minted}
\usepackage{ulem}
\usepackage{xcolor}
\usepackage{longtable}
\usepackage{makecell} % for multirowcell
\usepackage{listings}
\usepackage{xcolor}
\usepackage[T1]{fontenc}

\lstdefinestyle{yamlstyle}{
    basicstyle=\ttfamily\footnotesize,
    numbers=left,
    numberstyle=\tiny,
    stepnumber=1,
    numbersep=6pt,
    frame=single,
    framerule=0.5pt,
    breaklines=true,
    breakatwhitespace=true,
    tabsize=2,
    captionpos=b,
    keywordstyle=\color{blue},
    commentstyle=\color{gray},
    stringstyle=\color{teal},
}

\newlength\savewidth

\definecolor{baselinecolor}{HTML}{d6eaf8}

\definecolor{mygray}{gray}{0.4}

\AtBeginEnvironment{tcolorbox}{\tiny}

\newcount\Comments  %
\usepackage{color}
\definecolor{darkred}{rgb}{0.9,0,0}
\definecolor{darkgreen}{rgb}{0,0.5,0}
\definecolor{darkblue}{rgb}{0,0,0.7}
\definecolor{purple}{rgb}{.6, 0,.6}
\definecolor{orange}{rgb}{1.0,0.64,0}

\definecolor{deemph}{gray}{0.6}

\definecolor{baselinecolor}{gray}{.9}
\definecolor{yellow}{RGB}{218,165,32}
\definecolor{lightcyan}{rgb}{0.88, 1.0, 1.0}
\definecolor{lightskyblue}{rgb}{0.53, 0.81, 0.98}
\definecolor{aliceblue}{rgb}{0.94, 0.97, 1.0}
\definecolor{LightSlateBlue}{RGB}{70,130,180}
\definecolor{DeepBlue}{RGB}{65,100,170}
\definecolor{DeepPurple}{RGB}{136,105,160}
\definecolor{LightGreen}{RGB}{59,125,35}
\definecolor{LightRed}{RGB}{234,66,53}
\definecolor{cvprblue}{rgb}{0.21,0.49,0.74}
\newcommand{\grayrow}{\rowcolor[rgb]{0.95,0.95,0.95}}

\usepackage{shlab}
\usepackage[authoryear, sort&compress, round]{natbib}
\definecolor{cvprblue}{rgb}{0.21,0.49,0.74}
\title{ForceVLA2:
Unleashing Hybrid Force-Position Control with Force Awareness 
for Contact-Rich Manipulation}

\renewcommand{\today}{}
\author{
    Yang Li$^{2,1,9*}$ ,
    Zhaxizhuoma$^{3,1,9*}$,
    Hongru Jiang$^{3*}$, 
    Junjie Xia$^{1,9*}$\\
    Hongquan Zhang$^{6,4,1}$,
    Jinda Du$^{3}$,
    Yunsong Zhou$^{1\dag}$,
    Jia Zeng$^{1\dag}$,
    Ce Hao$^{7,8}$,
    Jieji Ren$^{3}$ \\
    Qiaojun Yu$^{1,3\dag}$,
    Cewu Lu$^{3,4,5\dag}$,
    Yu Qiao$^{1\dag}$,
    Jiangmiao Pang
}
\vspace{10pt}
\affil[*]{%
   Equal contributions
}
\affil[$^\dag$]{%
\mbox{\dag Corresponding authors (yuqiaojun@pjlab.org.cn)}
   \mbox{$^1$ Shanghai AI Laboratory \\$^2$ Tongji University\\$^3$Shanghai Jiao Tong University}

   \mbox{$^4$ Shanghai Innovation Institute\\$^5$ Noematrix Intelligence\\
$^6$ East China Normal University}

   $^7$ Zhongguancun Academy \\
   $^8$ National University of Singapore \\
   $^9$ Lumos Robotics

}

\begin{document}

% \input{preamble}
% \maketitle

% \twocolumn[{%
% \renewcommand\twocolumn[1][]{#1}%
% \maketitle
% \input{sec/0_figure}
% }]
% \vspace{-12pt}

% \vspace*{-2em}
% \vspace{-4cm}

\begin{abstract}

Embodied intelligence for contact-rich manipulation has predominantly relied on position control, while explicit awareness and regulation of interaction forces remain under-explored, limiting stability, precision, and robustness in real-world tasks. We propose ForceVLA2, an end-to-end vision--language--action framework that equips robots with hybrid force-position control and explicit force awareness. ForceVLA2 introduces force-based prompts into the VLM expert to construct force-aware task concepts across stages, and employs a Cross-Scale Mixture-of-Experts (MoE) in the action expert to adaptively fuse these concepts with real-time interaction forces for closed-loop hybrid force--position regulation. To support learning and evaluation, we construct ForceVLA2-Dataset, containing 1{,}000 trajectories over 5 contact-rich tasks, including wiping, pressing, and assembling, with multi-view images, task prompts, proprioceptive state, and force signals. Extensive experiments show that ForceVLA2 substantially improves success rates and reliability in contact-rich manipulation, outperforming $\boldsymbol{\pi}_0$ and $\boldsymbol{\pi}_{0.5}$ by 48.0\% and 35.0\%, respectively, across the 5 tasks, and mitigating common failure modes such as arm overload and unstable contact,thereby actively advancing force-aware interactive physical intelligence in VLAs. The project page is available at this \texttt{\href{https://sites.google.com/view/force-vla2/home}{URL}}.

\end{abstract}

\maketitle

\vspace{25pt}
\begin{center}
\centering
\captionsetup{type=figure}
\vspace{-15pt}
\includegraphics[width=0.95\textwidth]{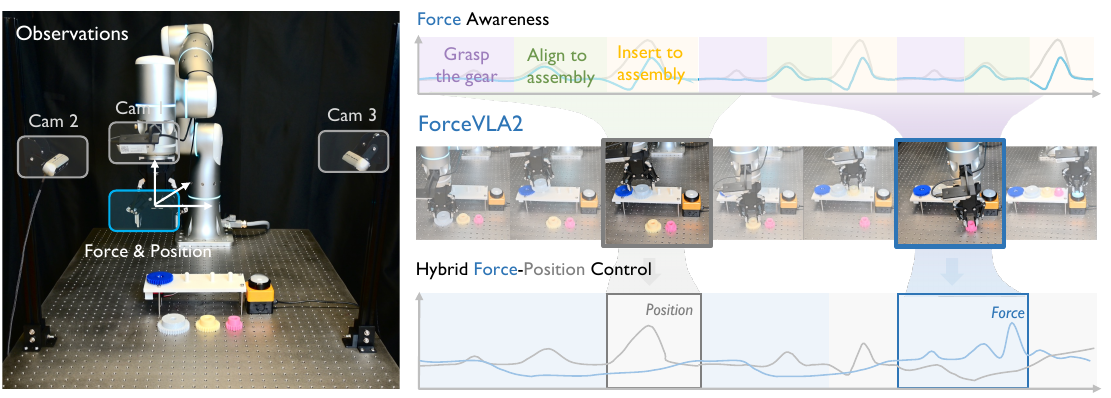}
\vspace{-8pt}
\captionof{figure}{\label{fig:teaser}
% \TR{
\textbf{ForceVLA2 concept.}
Contact-rich manipulation requires force regulation, beyond visual and state observations (left).
\textbf{ForceVLA2} integrates force information across multiple scales, enabling rich modeling of contact dynamics.
It builds force awareness into task planning through incoming force signals, and it outputs hybrid force–position actions with dynamic balance (right). 
%
% The left part shows the experimental setup and the vision-force observations for contact-rich manipulation.
% \textbf{Foca-VLA} realizes the deconstruction of complex tasks into force-aware sub-task prompts to construct task concepts, and implements a short-horizon reactive manipulation skill that adapts the regulation of hybrid force–position interaction in complex manipulations.
}
% }
\end{center}
\vspace{25pt}

\vspace{-7mm}

\section{Introduction}
\label{sec:intro}

Artificial intelligence has advanced rapidly with large vision–language models (VLMs)~\cite{zhang2024vision,yang2025qwen3,team2024qwen2,chen2024internvl,beyer2024paligemma, yuan2025embodied} that exhibit remarkable perceptual and reasoning capabilities. However, these models remain confined to virtual domains, lacking the embodiment necessary for authentic physical understanding and interaction in real-world settings. To overcome this limitation, vision–language–action models (VLAs)~\cite{bjorck2025gr00t,black2410pi0,kim2024openvla, zhao2025cot, zheng2025tracevlavisualtraceprompting} extend VLMs toward physical intelligence by seamlessly connecting perception and reasoning to embodied interaction. This integration unifies cognition and manipulation into a cohesive framework, enabling robots not only to perceive and understand, but also to manipulate objects in the physical world in ways that closely mirror human intelligence. Consequently, VLAs have become a foundation for robotic learning, paving the way for enhanced sample efficiency, generalizability, compositionality, and incremental learning~\cite{kaelbling2020foundation}, and enabling adaptive, context-aware behavior in dynamic, complex environments.

While recent VLAs such as OpenVLA~\cite{kim2024openvla}, OpenVLA-OFT~\cite{kim2025fine}, GR00T-N1~\cite{bjorck2025gr00t}, and $\boldsymbol{\pi}_0$~\cite{black2410pi0} have advanced embodied cognition by linking perception and language to action, these models demonstrate strong performance in semantic grounding and language following, showing that pretrained multimodal models can acquire priors and be efficiently fine-tuned with only a few demonstrations. However, current VLAs still lack the ability to reason about physical dynamics and fine-grained contact interactions, which are essential for real-world, contact-rich manipulation. Building upon $\boldsymbol{\pi}_0$~\cite{black2410pi0}, ForceVLA~\cite{yu2025forcevla} extends this framework by incorporating the force modality, leading to improved success rates in contact-rich manipulation tasks.

Yet, existing VLAs reduce force to an auxiliary perceptual input rather than leveraging it for active, adaptive closed-loop force interaction, a core requirement for genuine physical intelligence. In contrast, humans rely on high-level visual and linguistic reasoning to identify stage-specific targets and estimate coarse spatial relations, while integrating real-time force sensing as direct physical perception to adaptively refine both force and position during interaction. Current VLAs, however, still lack mechanisms to reason about the spatiotemporal relationships required across different task stages and to integrate force perception with active force–position interaction within subtasks. This gap motivates a unified treatment of \textit{force awareness} and \textit{closed-loop force–position control} within the VLA paradigm for contact-rich manipulation.

To overcome these limitations, we propose \texttt{\textbf{ForceVLA2}}, a novel framework that equips VLAs with active hybrid force-position control with force awareness to enhance contact-rich manipulation, as shown in~\cref{fig:teaser}. By incorporating force modality into the VLA architecture, ForceVLA2 links language-level and visual semantic reasoning with physical contact-level interaction, enabling physically grounded, generalizable, and highly adaptive behaviors, especially in tasks requiring fast and stable force-following. 
Inspired by the human motor control system, where force perception and regulation are hierarchically integrated~\cite{hogan2018impedance}, ForceVLA2 employs a dual-level design.
A long-horizon reasoning module driven by a force prompt is embedded into the VLM expert to construct force-aware task concepts, while a short-horizon reactive manipulation skill employs a Cross-Scale  Mixture-of-Experts (MoE) to integrate force-aware task knowledge from the VLM expert with embodied interaction forces in the action expert, enabling adaptive regulation of hybrid force–position interaction. Extensive experiments demonstrate that ForceVLA2 consistently improves performance and generalization in contact-rich manipulation. Our contributions are summarized as follows:

\begin{itemize}
    \item We introduce ForceVLA2, the first end-to-end hybrid force–position control framework with force awareness for VLAs, enhancing contact-rich manipulation by integrating force-prompt-driven VLM reasoning and fusing force into action experts via a Cross-scale MoE, enabling coordinated hybrid force–position interaction and tightly linking VLM reasoning with action generation to improve physical interaction.
    \item We construct ForceVLA2-Dataset, a dataset of 1,000 trajectories across 5 contact-rich tasks, including wiping, pressing, and assembling, with fundamental modalities such as multi-view images, task prompts, and proprioceptive states. It is further enhanced by adding force prompts and incorporating force into the action space, bridging the gap between embodied hybrid force–position interaction and force-aware task concepts, providing an empirical foundation for advancing research from force perception to interaction.
    \item We conducted extensive experiments demonstrating the effectiveness of ForceVLA2, compared to $\boldsymbol{\pi}_0$ and $\boldsymbol{\pi}_{0.5}$, the success rates increased by 48.0\% and 35.0\%, respectively, across the 5 tasks, represents notable progress toward bridging force-aware task concept with embodied adaptive hybrid force–position manipulation skills.

\end{itemize}

\section{Related Work}
\label{sec:related}

\noindent \textbf{Vision-language-action models.}
Recent research on VLAs~\cite{rt1, brohan2023rt, kim2024openvla, kim2025fine, vlas, robomamba, vima, interleave, starvla2025} leverages large-scale multimodal pretraining to generalize robotic policies across diverse tasks and embodiments, mapping visual observations and language instructions to low-level actions through end-to-end learning.
Autoregressive models~\cite{kim2024openvla, kim2025fine, pertsch2025fast} treat control as next-token prediction over discretized action tokens, enabling VLM-style training with efficient decoding.
Diffusion-based approaches~\cite{team2024octo, dp, prediction, chatvla, hybridvla} model continuous action distributions for more diverse and controllable behaviors, though at high computational cost.
Flow-based methods such as $\pi_0$~\cite{black2410pi0} accelerate action generation via flow-matching decoders, while ForceVLA~\cite{yu2025forcevla} extends it with force inputs to improve contact-rich manipulation accuracy.
Dual-system designs like $\pi_{0.5}$ and GR00T-N1~\cite{bjorck2025gr00t} separate high-level reasoning from low-level action generation, combining generalization with action.
% Yet, current VLAs largely omit force sensing in observations and lack active force control in action space, limiting their adaptability to contact-rich manipulation.

\noindent \textbf{Physical modality integration in robot policies.}
% 
% Traditional vision-only robotic manipulation methods struggle with dynamic interactions that require fine-grained feedback. 
% Thus, recent research has focused on integrating force and tactile sensing into robotic control systems. 
% Recent advancements in robotic manipulation have increasingly recognized the importance of physical modalities (\textit{e.g.}, force and tactile) in contact-rich tasks.
% Studies such as TacDiffusion~\cite{tacdiffusion}, AdaptiveCP~\cite{adaptiveCP}, ForceMimic~\cite{forcemimic}, and FoAR~\cite{force-aware} highlight the benefits of force feedback for enhancing manipulation performance. 
% 
Traditional VLA models struggle with dynamic, contact-rich interactions that demand fine-grained feedback, motivating recent efforts to integrate force and tactile modalities for improved control.
Recent works incorporating force sensing~\cite{tacdiffusion,adaptiveCP,forcemimic,force-aware,yu2025forcevla} enhance motion stability and contact precision, yet force is typically used only as a perceptual cue rather than an active control signal.
Tactile-based methods such as TLA~\cite{tla}, Tac-Man~\cite{tac-man}, and other multimodal fusion approaches~\cite{impact,should, Huang2025TactileVLA} improve robustness under occlusion but cannot substitute for direct force sensing, since tactile-to-force estimation remains indirect and noisy. Despite these advances, existing methods remain limited to force perception and purely position-based interaction. 
% Foca-VLA advances beyond this limitation with active hybrid force–position control, enabling robots to actively sense and regulate contact forces for stable, fine-grained manipulation.

\noindent \textbf{Multimodal embodied datasets.}
Large-scale embodied datasets have significantly advanced robot learning, yet they rarely include standardized force modality, which is essential for modeling contact~\cite{wu2025robomind, AgiBotWorldTeam2025agibot-world-colosseo, walke2024bridgedatav2datasetrobot}.
Open-X-Embodiment~\cite{open_x_embodiment_rt_x_2023} aggregates over one million trajectories across 22 robots, though only a subset of its datasets include force modalities, leading to sparse and inconsistent coverage.
Recent datasets have begun to incorporate explicit force modality for contact-rich manipulation. RH20T~\cite{fang2024rh20t} provides multi-modal data, including RGB-D, audio, proprioceptive state, and calibrated force collected via haptic teleoperation, while REASSEMBLE~\cite{sliwowski2025reassemblemultimodaldatasetcontactrich} focuses on assembly tasks with synchronized visual, proprioceptive, and force modalities, incorporating sub-tasks to enable detailed analysis of manipulation skills.
% Despite this progress, prior datasets largely treat force as observation only. Our dataset advances beyond visual and physical perception, further augmented with stage-specific force prompts that enable an explicit hybrid force–position action space for policy learning.

% Despite the success of force and tactile sensing in robotic manipulation, existing methods often rely on static sensory fusion techniques, where the different modalities are merged at predetermined stages. This static fusion limits the model's ability to adapt to dynamic changes in the environment. Moreover, few studies have explored dynamic routing mechanisms that enable the seamless integration of force, tactile, and visual modalities in real-time decision-making. The lack of a fully unified multimodal framework restricts the potential for robots to perform complex tasks in unstructured, real-world environments.

% Thus, while significant strides have been made in integrating physical modalities into robotic control, challenges remain in developing fully unified models that dynamically fuse vision, language, and tactile/force feedback. Addressing these challenges will be crucial for enabling robots to perform robust, contact-rich tasks with high precision and adaptability in diverse, real-world scenarios.

\section{ForceVLA2 Framework}

\begin{figure*}[ht]
\centering
\includegraphics[width=0.85\textwidth]{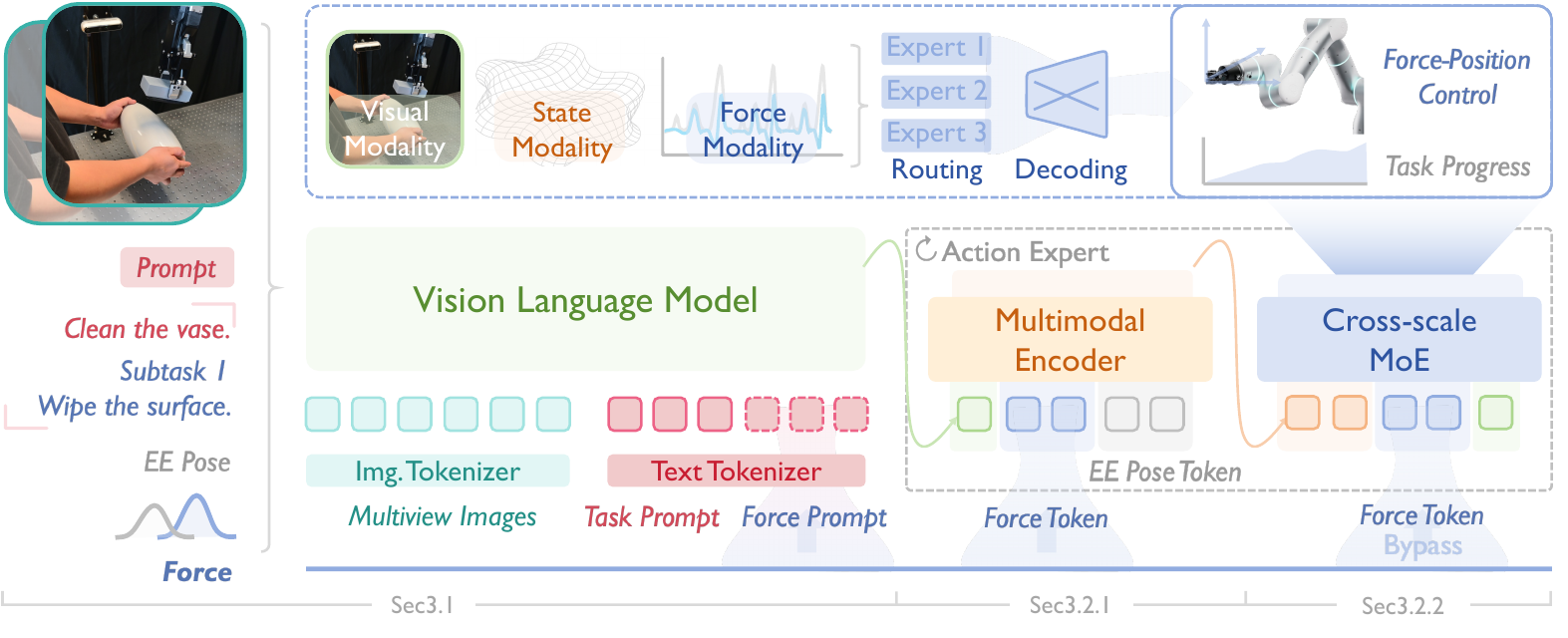}
\caption{
\textbf{Framework of
ForceVLA2.} 
% \TR{
ForceVLA2 takes multi-view images, task and force prompts, and proprioceptive states (EE pose and force) as input.
Force is injected at multiple scales: as sub-task prompts fused with images in the vision-language model, and as force tokens combined with EE pose in the multimodal encoder, with a bypass to preserve raw signals.
The cross-scale MoE integrates these modalities to produce hybrid force–position actions and track sub-task progress for adaptive, contact-rich manipulation.
% 
% The VLM expert,  capable of long-horizon reasoning driven by multiview images, task prompts, and force prompts, constructs force-aware task concepts. Meanwhile, the action expert, with a short-horizon reactive manipulation skill, employs a cross-scale routing MoE to integrate force-aware task knowledge from the VLM expert with embodied EE 6D poses and interaction forces, enabling adaptive hybrid force–position interaction.
}
% }
\label{fig:pipeline}
\vspace{-4mm}
\end{figure*}

ForceVLA2 is designed to unify force awareness and hybrid force–position control within a vision–language–action (VLA) framework, thereby enabling precise, adaptive, and stable contact-rich manipulation. Inspired by findings in human sensorimotor control, we posit that force acts as a unifying signal across the perception–planning–execution hierarchy, providing essential feedback for dynamic and compliant interaction~\cite{hogan2018impedance,wolpert1995internal}. Neuroscientific evidence further supports this perspective, showing that humans perceive external forces through proprioceptive sensing and regulate them via internal forward models implemented in the cerebellum and motor cortex~\cite{hogan2018impedance}. 
Motivated by biological sensing–acting mechanisms, ForceVLA2 internalizes multi-scale force cues within both context-dependent long-horizon reasoning, realized through a force prompt, and adaptive, stage-aware reactive skills, enabling precise force–position interaction for unified, force-aware manipulation that bridges perception and action.

Building upon these principles, we propose the ForceVLA2 architecture, which integrates multi-scale perception, contextual reasoning, and force-aware manipulation into a unified VLA framework. 
As shown in~\cref{fig:pipeline}, multi-view images, task prompts, and force prompts are jointly encoded within a VLM, providing high-level guidance for task operation sequencing and spatial–semantic reasoning in task decomposition (\cref{sec:long-horizon}). 
In parallel, force observation bypasses high-level fusion and modulates the action expert via a direct gradient pathway, enabling a reactive response to observed force during contact (\cref{sec:encoding}).
Force and proprioceptive tokens are routed through a Cross-Scale Mixture-of-Experts (MoE)~\cite{shazeer2017outrageously}, which dynamically integrates visual, language, and force information to determine the dominant modality at each phase of the task (\cref{sec:fusion and decoding}). ForceVLA2 unifies force–position control and task progress prediction into a cohesive, force-aware closed-loop architecture, enabling physically grounded reasoning and adaptive perception–action coordination for complex contact-rich manipulation.

\subsection{Long-Horizon Force Awareness via Prompting}
\label{sec:long-horizon}
Current VLA models rely on pre-trained priors for task decomposition; however, purely vision–language inputs are insufficient in contact-rich settings that require human-like task awareness and physically grounded reasoning about task concepts, particularly interaction forces~\cite{din2025vision}.
In such contact-rich scenarios, visual cues are often ambiguous or insufficient, whereas essential information is conveyed through force feedback, as in aerospace connector insertion or gear assembly.
To address this, we introduce force prompts as textual cues that indicate the current subtask and encode stage-specific physical context, thereby constructing force-aware task concepts.
This mechanism enables ForceVLA2 to inherit VLM knowledge, assess subtask completion, transition across stages, and explicitly update force cues to guide force-aware manipulation.

\noindent
\textbf{Vision, task, and force prompt integration.}
Inspired by the $\pi_0$ framework~\cite{black2410pi0}, the ForceVLA2 architecture integrates vision, task prompts, and force prompts before a conditional flow-matching action expert.
Visual observations from multiple RGB cameras, along with task instructions, are encoded by a SigLIP-based~\cite{siglip} vision–language model (based on PaliGemma~\cite{beyer2024paligemma}) into contextual embeddings.
The text prompt provides a global description of the task, while the force prompt encodes the current subtask state. 
Each task has a predefined list of subtasks, and the force prompt determines whether to maintain the current subtask or transition to the next, functioning like a discrete-state machine.

Specifically, the visual tokens $\mathbf{E}_v \in \mathbb{R}^{N_v \times D_\text{model}}$ 
are first processed through a visual encoder $f(\cdot)$: $\mathbf{Z}_v = f(\mathbf{E}_v) \in \mathbb{R}^{N_v \times D_\text{model}}$.
The linguistic tokens, obtained by concatenating the text prompt $\mathbf{T}_t \in \mathbb{R}^{N_t \times D_\text{model}}$ and the force prompt $\mathbf{T}_f \in \mathbb{R}^{N_f \times D_\text{model}}$, 
are processed through a text encoder $g(\cdot)$: $\mathbf{Z}_l = g([\mathbf{T}_t; \mathbf{T}_f]) \in \mathbb{R}^{(N_t+N_f) \times D_\text{model}}$,
where $N_t$ and $N_f$ are the number of text and force prompt tokens.
The outputs of the visual and text encoders are concatenated to form the input to the vision–language model: $\mathbf{E}_\text{in} = [\mathbf{Z}_v; \mathbf{Z}_l] \in \mathbb{R}^{(N_v + N_t + N_f) \times D_\text{model}}$.
Finally, $\mathbf{E}_\text{in}$ undergoes token-to-token attentions within the vision–language model ($\texttt{VLM}(\cdot)$), 
producing the fused contextual tokens:
\begin{equation}
    \mathbf{E} = \texttt{VLM}(\mathbf{E}_\text{in}) \in \mathbb{R}^{(N_v + N_t + N_f) \times D_\text{model}}.
\end{equation}
The resulting sequence $\mathbf{E}$ serves as the fused multi-modal representation, 
which is subsequently used for downstream short-term force perception and control.

% \subsection{Delicate Manip. via Force-Position Control}
\subsection{Short-Horizon Force-to-Control Loop}
\label{sec:short-horizon}

% Existing VLA models rely on position control, predicting end-effector poses from visual and state observations.
% This design suits coarse manipulation, where gripper–object interactions are visually salient.
% Yet, in delicate manipulation, subtle visual cues make pure position control insufficient for precise and adaptive contact handling.
% To address this, we incorporate transient force awareness and force control into the model, forming a force-feedback loop that enables hybrid force–position control for contact-rich tasks.

Existing VLA models rely on end-effector (EE) 6D pose control, which suffices for coarse tasks but fails when subtle force interactions dominate.
ForceVLA2 encodes proprioceptive state and force differently within the action expert, passing them through the multi-modal encoder to fuse with visual–language embeddings for long-horizon reasoning, while transient force signals bypass the fusion layer and feed directly to the MoE, forming a short-term reactive loop.
This design preserves gradient fidelity for rapid force feedback, avoids over-reliance on past trajectories, and enables active exploration (Appendix {\color{red}A}).
The MoE dynamically integrates long- and short-term cues to produce joint predictions of EE 6D pose, force, and subtask progress for hybrid force-position control in contact-rich manipulation.

\subsubsection{multi-modal Encoding for EE 6D pose and Force}
\label{sec:encoding}

\noindent
\textbf{Proprioceptive-force encoding.}
% This component builds upon the Action Expert module in the $\pi_0$ framework, which employs a flow-matching mechanism for action generation.
The proprioceptive state, represented by the EE 6D pose, is expressed as
$\mathbf{p} \in \mathbb{R}^7$
and encoded through another linear layer $\phi_P$:
\begin{equation}
    \mathbf{E}_P = \phi_P(\mathbf{p}) \in \mathbb{R}^{D_\text{model}}.
\end{equation}

To enable force-aware perception, the raw 6D force–torque readings
$\mathbf{f}_{\text{raw}} \in \mathbb{R}^6$
are first projected into the model embedding space via a linear mapping $\phi_F$:
\begin{equation}
    \mathbf{E}_F = \phi_F(\mathbf{f}_{\text{raw}}) \in \mathbb{R}^{D_\text{model}}.
\end{equation}

The encoded EE 6D pose and force tokens are concatenated to form a multi-modal state representation,
$\mathbf{E}_\text{state} = [\mathbf{E}_P; \mathbf{E}_F]$,
which is then fed into the MoE alongside visual–language contextual tokens $\mathbf{E} \in \mathbb{R}^{N_\text{VL} \times D_\text{model}}.$
% The position embedding $\mathbf{E}_P$ is combined with the visual–language contextual tokens 
% $\mathbf{E} \in \mathbb{R}^{N_\text{VL} \times D_\text{model}}$ 
% and fed into the Action Expert for cross-modal fusion and long-horizon reasoning. 
Meanwhile, a force embedding $\mathbf{E}_F$ bypasses the cross-modal interaction and is sent directly to the MoE, forming a short-term, reactive pathway. 
% Together, these pathways enable the model to integrate long-term positional guidance with fast force feedback.

\noindent
\textbf{Cross-modal interaction.}
To incorporate high-level task context into proprioceptive sensing, the state embeddings interact with the vision–language features through cross-attention. 
Specifically, the encoded visual–language sequence $\mathbf{E} \in \mathbb{R}^{N_\text{VL} \times D_\text{model}}$ serves as the contextual key–value pair, while the state tokens act as queries:
\begin{equation}
    \mathbf{E}'_{\text{state}} = \texttt{CrossAttn}(\mathbf{E}_{\text{state}}, \mathbf{E}).
\end{equation}
This operation injects global task semantics into the local proprioceptive streams, allowing the model to interpret force and motion signals under the guidance of visual–language intent.
The conditioned embeddings are then concatenated as:
\begin{equation}
    \mathbf{E}_\text{cond} = [\mathbf{E}; \mathbf{E}'_{\text{state}}; \mathbf{E}_F],
\end{equation}
which forms the multi-modal input for the following expert routing stage.

\subsubsection{Adaptive Routing and Decoding}
\label{sec:fusion and decoding}

\noindent
\textbf{Cross-scale Mixture-of-Experts.}
The conditioned sequence $\mathbf{E}_\text{cond}$ is processed by a sparse MoE layer designed to adaptively select the most informative modality for the current manipulation phase.
The MoE contains three modality-specific experts (visual, state, and force), each implemented as a lightweight MLP~\cite{haykin1994neural}.
A dynamic gating network computes token-wise routing weights
$\mathbf{w} = [w_V, w_S, w_F]$,
and activates the most relevant expert for each token:
\begin{equation}
    \mathbf{E}_\text{MoE} = \sum_{m \in \{V,S,F\}} w_m \cdot \texttt{Expert}_m(\mathbf{E}_\text{cond}).
\end{equation}
Through this mechanism, the network autonomously emphasizes visual reasoning during free-space motion and force or position cues when contact and compliance become critical.
The fused representation $\mathbf{E}_\text{MoE}$ is then projected to the latent space of the flow-matching policy~\cite{flowmatching} for downstream action generation.

\noindent
\textbf{Flow Matching policy for position-force control.}
The flow-matching policy head generates force-aware actions by progressively denoising a noise-initialized action conditioned on the fused multi-modal context.
At the beginning of each time step $t$, a noisy action sample 
$\mathbf{a}_t^{(0)} \sim \mathcal{N}(0, I)$
is iteratively refined according to the learned conditional flow:
\begin{equation}
    \frac{d\mathbf{a}_t^{(\tau)}}{d\tau}
    = F_\theta\big(\mathbf{a}_t^{(\tau)}, \mathbf{E}_\text{MoE}, \tau \big),
     \tau \in [0, 1],
\end{equation}
where $\tau$ is the denoising time, $\mathbf{E}_\text{MoE}$ provides the contextual multi-modal embedding from the MoE module.
In practice, this continuous flow is implemented as a discrete denoising process:
\begin{equation}
    \mathbf{a}_t^{(\tau+1)} = 
    \mathbf{a}_t^{(\tau)} + 
    \Delta_\tau \cdot F_\theta\big(\mathbf{a}_t^{(\tau)}, \mathbf{s}_t, \mathbf{E}_\text{MoE}\big),
\end{equation}
starting from random noise and gradually converging to the final force–position command:
\begin{equation}
    \mathbf{a}_t = \mathbf{a}_t^{(1)} = [\Delta \mathbf{p}_t; \mathbf{f}_t], \Delta \mathbf{p}_t \in \mathbb{R}^7,~\mathbf{f}_t \in \mathbb{R}^6,~s_t \in [0,1].
\end{equation}
Here, $\Delta \mathbf{p}_t$ represents the desired change in end-effector pose,
$\mathbf{f}_t$ denotes the predicted contact force.
% , and $s_t$ is a scalar transition indicator that signals subtask completion.
By conditioning the denoising process on the fused visual–language–force representation, the model achieves closed-loop, context-aware control that adapts fluidly to contact-rich interaction.

\noindent
\textbf{Probabilistic modeling of force-control transition.}
Note that the transition indicator $s_t$ is modeled as a continuous floating-point value output by the model, ranging from 0 to 1. 
Once $s_t$ exceeds 1 (threshold), the controller transitions to the next subtask and resets the value to 0. 
The ground truth value for $s_t$ during training, denoted as $\hat{s}_t$, is computed based on the observed quantities.
Let $A$ denote the current end-effector position and $A_t$ the target position, $E$ be the current end-effector orientation,  $E_t$ the target orientation, and $\overrightarrow{AA_t}$ represent the displacement from $A$ to $A_t$. Define the observed statistics:
$\Theta = \tfrac{1}{2}\left(\frac{\overrightarrow{E} \cdot \overrightarrow{E_t}}{\lVert \overrightarrow{E} \rVert \lVert \overrightarrow{E_t} \rVert} + 1\right) \in (0, 1)$,
$L = \lVert \overrightarrow{AA_t} \rVert \in (0, \infty)$,
$F = \lVert \mathbf{f} \rVert \in [n, m]$.
Here, $\Theta$ measures orientational alignment (1 = aligned, 0 = opposite), $L$ is the remaining distance to the goal, and $F$ represents the instantaneous contact force (bounded within the normalization limits $n$ and $m$).

We define the ground truth for the subtask transition probability as the joint event that these quantities satisfy their respective progress conditions. Given observed values $(\theta, l, f)$ and assuming
$\Theta \sim \texttt{Beta}(\alpha, 1)$,
$L \sim \texttt{Exponential}(\lambda)$,
$F \sim \texttt{Uniform}(n, m)$,
we calculate the subtask transition probability $\hat{s}_t$ as follows:
\begin{align}
\hat{s}_t \; &= P(\Theta \le \theta,\; L \ge l,\; F \le f) \notag\\ &= \int_0^{\theta}\!\frac{\Gamma(\alpha+1)}{\Gamma(\alpha)}\,\theta'^{\,\alpha-1} d\theta' \!\cdot\! \int_l^{\infty}\!\lambda e^{-\lambda \ell}\, d\ell \!\cdot\! \int_n^{f}\!\frac{1}{m-n}\,df \notag\\ &= \frac{\Gamma(\alpha+1)}{\alpha\Gamma(\alpha)}\,\theta^{\alpha} e^{-\lambda l}\,\frac{f-n}{m-n}.
\end{align}
Here, $\Gamma(\cdot)$ represents the Gamma function, with $\alpha = 2$ and $\lambda = 2$. $s_t$ serves as an indicator of subtask completion. When it exceeds a threshold, the controller transitions to the next subtask and resets $s_t$ to 0.

\begin{figure*}[t]
\centering
\includegraphics[width=\textwidth]{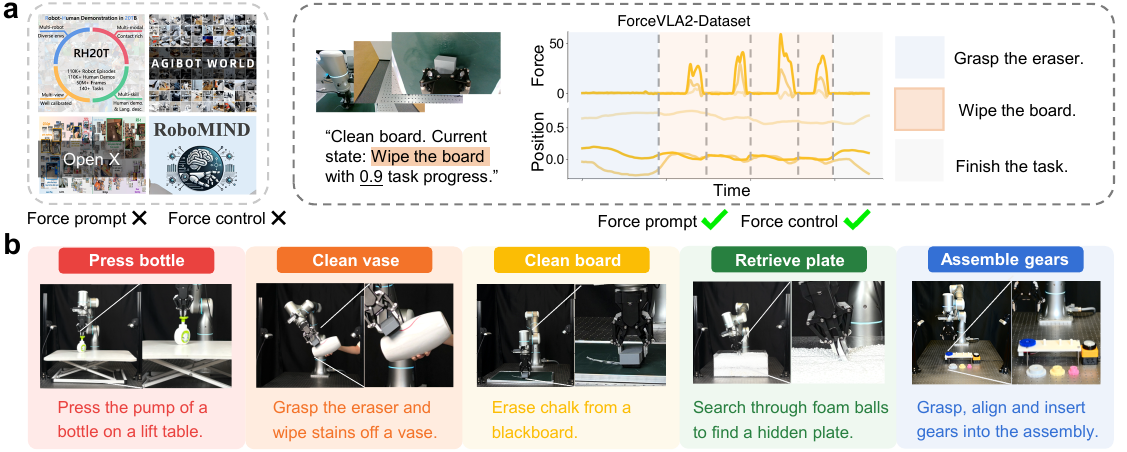}
\caption{
% \TR{
\textbf{The illustration of ForceVLA2-Dataset.}
(a) ForceVLA2-Dataset is the first dataset with force prompts for task decomposition and the only one providing force-control supervision. (b) It features 1,000 demonstrations across five contact-rich tasks.
% : \textsc{PRESS Bottle}, \textsc{Clean Vase}, \textsc{Wipe Board}, \textsc{Retrieve Plate}, and \textsc{Assemble Gears}.
}
\label{Fig: Dataset placeholder}
\vspace{-5mm}
\end{figure*}

\section{ForceVLA2-Dataset}

To train the proposed ForceVLA2 model, we collected a new dataset specifically tailored for contact-rich manipulation tasks, with a particular emphasis on capturing force-control signals for sub-task segmentation and annotation (\cref{Fig: Dataset placeholder} (a)).
Unlike previous work, which primarily relied on EE 6D pose trajectories~\cite{open_x_embodiment_rt_x_2023,fang2024rh20t} and VR-based human teleoperation~\cite{xue2025reactive}, our setup employs the force-feedback GELLO teleoperation framework~\cite{wu2023gello} to capture high-quality demonstrations with natural motion dynamics, while explicitly recording force and motion trajectories.
Each manipulation task is segmented into 3–5 annotated subtasks, guided by visual inspection and analysis of the synchronized force signals.
The collected dataset constitutes a multi-modal corpus encompassing visual, proprioceptive, task-prompt, force-prompt, and force modalities.

\begin{figure}[t]
    \centering
    \includegraphics[width=0.6\linewidth]{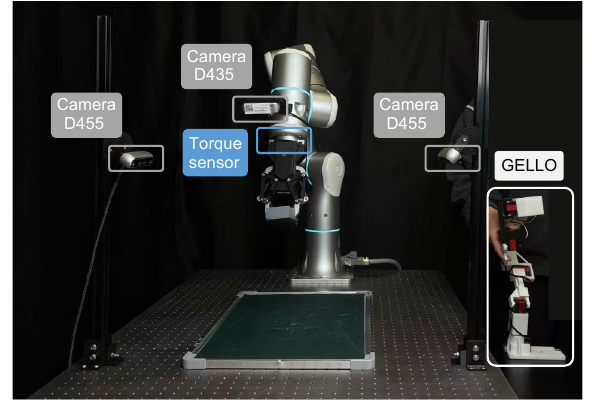}
    \caption{
    % \TR{
    \textbf{The dataset collection system.} A Flexiv arm is driven by manually controlled GELLO~\cite{wu2023gello} to accomplish dexterous tasks and record images, force, as well as the pose of the robot.
    % }
    }
    \label{fig:datasetCap}
\vspace{-6mm}
\end{figure}

\noindent
\textbf{Hardware and sensor setup.}
Demonstrations were collected using a 7-DOF Flexiv Rizon 4s robotic arm equipped with a DH Robotics AG-95 adaptive gripper. Visual observations were acquired from three RGB cameras: two static third-person views (Intel RealSense D455, 1280×720 @ 30 fps) and one wrist-mounted egocentric view (Intel RealSense D435, 640×480 @ 30 fps), as shown in~\cref{fig:datasetCap}. A 6D force/torque sensor attached to the end-effector recorded interaction forces at 300 Hz, while the robot joint states and end-effector (EE) 6D poses were logged synchronously. All visual streams were resized to 480×640 resolution, normalized, and timestamp-synchronized prior to storage.

\noindent
\textbf{Collected tasks.}
We curated 5 contact-rich manipulation tasks: \textit{press bottle}, \textit{clean vase}, \textit{clean board}, \textit{retrieve plate}, and \textit{assemble gears} (\cref{Fig: Dataset placeholder} (b)).
Operators used the force-feedback GELLO teleoperation interface to control the robot in real time, ensuring accurate and repeatable interaction patterns while preserving natural force application.
To enhance diversity, object poses are varied across demonstrations.
Each task is recorded until completion, and subtask boundaries are annotated offline based on distinct transitions in the force signal profiles.

\noindent
\textbf{Dataset statistics.}
The resulting dataset comprises 1000 trajectories and approximately 500K synchronized timesteps.
Each trajectory contains time-aligned streams of RGB images, end-effector (EE) 6D poses, and force signals.
The ForceVLA2-Dataset facilitates research on force-aware and subtask-structured manipulation policy learning by introducing force prompts and integrating force into the action space, bridging the gap for embodied hybrid force–position interaction and providing an empirical foundation for advancing research from force perception to interaction.
The detailed dataset statistics are
 in the Appendix~{\color{red}B}.

% \newpage

%%%script%%%
% We take the simulation data to model a probability function $P(\text{switch to force control)}$ with three random variables $\Theta$, $L$, and $F$, where we denote the current and target position as $A$ and $B$, and let $\overrightarrow{e_{u}}$ be the unit basis vector in the target direction without loss of generality. And we shall define the following
% \begin{equation*}
%     \Theta = \frac{1}{2} \left( \frac{\overrightarrow{AB} \cdot \overrightarrow{e_{u}}}{\lVert \overrightarrow{AB} \rVert} +1 \right) \in (0,1),
% \end{equation*}
% $L \coloneqq \lVert \overrightarrow{AB} \rVert\in (0,+\infty)$ is the distance from $A$ to $B$, and $F \in [n,m]$ is the magnitude of the force. The three random variables $\Theta$, $L$, and $F$ follow $Beta(\alpha,1)$, $Exponential(\lambda)$ and $Uniform(n,m)$ respectively.
% Since they are pairwise independent, the probability function can be determined separately and multiplied together
% \begin{align*}
%     P &= P(\Theta \leq \theta, L \geq l, F\leq f), \\
%       &= P(\Theta \leq \theta) \cdot P(L \geq l) \cdot P( F\leq f), \\
%       &= \int_{0}^{\theta} \frac{\Gamma(\alpha+1)}{\Gamma(\alpha)\Gamma(1)} \theta^{\alpha-1} d\theta \int_{l}^{+\infty}  \lambda e^{-\lambda l} dl \int_{n}^{f} \frac{1}{m-n} df, \\
%       &= \frac{\Gamma(\alpha+1)}{\alpha\Gamma(\alpha)}\theta^{\alpha}e^{-\lambda l}\frac{f-n}{m-n}.
% \end{align*}

\section{Experiments}
In this section, we evaluate the effectiveness of ForceVLA2 on diverse contact-rich manipulation tasks that test perceptual and reasoning abilities, as well as adaptive, closed-loop hybrid force–position interactions across multiple task stages under complex physical conditions. The experiments address the following research questions:
\vspace{-5pt}
\begin{itemize}
    \item \textbf{Q1:}  How does ForceVLA2 perform in real-world contact-rich manipulation tasks, and what advantages does hybrid force–position control offer over pure position control?
    \item \textbf{Q2:} Which components contribute most significantly to the performance improvements of ForceVLA2?  

    \item \textbf{Q3:} Which modality fusion strategy yields the best performance in active hybrid force–position controlling?

\end{itemize}

\begin{figure*}
    \centering
    \includegraphics[width=\textwidth]{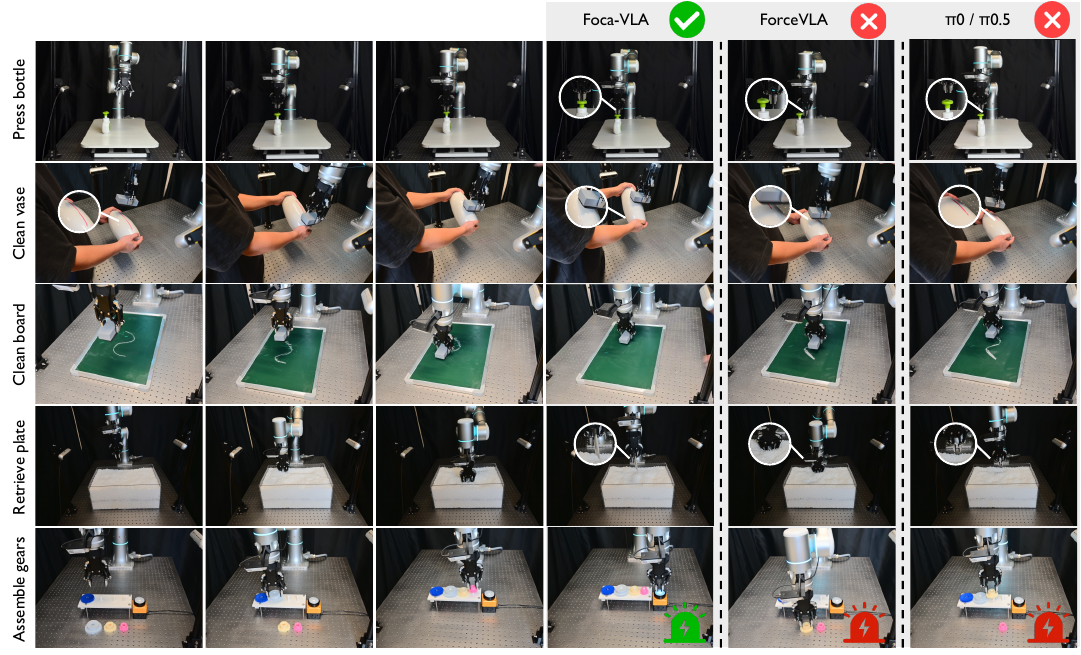}
    \caption{\textbf{Qualitative results on typical manipulation tasks (compared with ForceVLA and $\pi$ serials)}. ForceVLA2 completes these tasks with higher success rates and faster execution while avoiding arm overload, demonstrating superior compliance.}
    \label{Fig: main exp}
% \vspace{-1mm}
\end{figure*}

\subsection{Experiment Setting}
\noindent
\textbf{Tasks and evaluation metrics.}  Our experimental benchmark consists of 5 contact-rich manipulation tasks within the proposed ForceVLA2-dataset: Press the bottle, Clean the vase, Clean the board, Retrieve the plate, and Assemble gears. The primary evaluation metric is the success rate ($\%$), determined by conducting 20 independent trials for each task.

\noindent
\textbf{Baselines.} To demonstrate the superiority of our proposed ForceVLA2, we compare it against five established methods: ACP~\cite{adaptiveCP},  $\pi_0$~\cite{black2410pi0}, $\pi_{0.5}$~\cite{pi0.5}, ForceVLA~\cite{yu2025forcevla} and $\pi_0$ with naive force input. ACP leverages principles of admittance control to implicitly incorporate the desired force into the final output by predicting a virtual action. The $\pi_0$ model leverages a flow matching architecture integrated with a pre-trained VLM to leverage Internet-scale knowledge. Building upon $\pi_0$, the $\pi_{0.5}$ model incorporates co-training on diverse tasks to enhance generalization. 
% Two additional $\pi_0$-based baselines are implemented for comparison: one incorporating naive force-input injection ($\pi_0$ w/ F) and the other employing a low-stiffness impedance controller ($\pi_0$ w/ I).  
Meanwhile, ForceVLA prioritizes external force sensing as a core modality within its VLA architecture.

\begin{figure*}
    \centering
    \includegraphics[width=\textwidth]{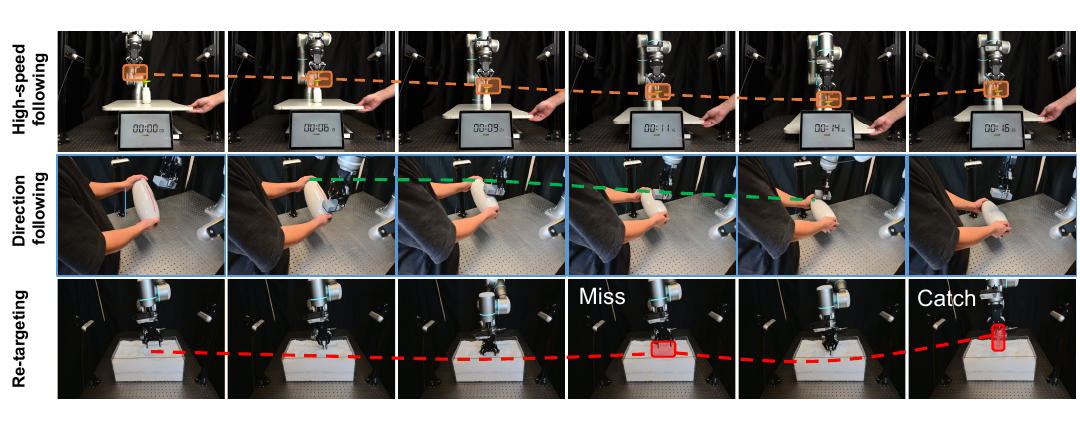}
    \caption{\textbf{Additional tests on following and re-targeting.} ForceVLA2 exhibits robust position and orientation following, and in object search tasks, it can still perform successful re-grasps even when visual observations fail.
}
\label{Fig: additional exp}
\vspace{-3mm}
\end{figure*}

\subsection{Main Experiment Results}
\vspace{-2mm}

\noindent\textbf{Overall performance on contact-rich tasks (Q1).}
As summarized in~\cref{main results}, ForceVLA2 significantly outperforms all baselines, achieving a 66$\%$ average success rate across all tasks. On force-sensitive tasks, it surpasses the second-best model by up to 50$\%$. Compared with models without force inputs, ForceVLA2 and ForceVLA, which incorporate force feedback, show remarkable improvements on force-sensitive tasks such as object search, demonstrating superior force perception. Meanwhile, $\pi_0$ w/ F reaches 17$\%$, which is similar to the base version, and prior analyzes (TA-VLA, ForceVLA) also suggest that simple concatenation can act as nuisance input and degrade pre-trained VLA representations, instead of providing consistent control-relevant signals. The ACP method achieves a success rate of only 16.0\%, primarily due to its limited generalization capabilities. 

\noindent\textbf{Force-intensive tasks and overload avoidance.}
As illustrated in~\cref{Fig: main exp}, for force-intensive tasks such as clean board and bottle pressing, excessive contact force can easily push the robot into overload. Among the compared methods, ForceVLA2 actively adjusts its interaction forces and successfully completes these tasks, whereas other VLAs lack sufficient reactive adjustment capabilities, leading to task failures.

% \begin{table}
% \centering
% \begin{tblr}{
%   cell{2}{1} = {r=2}{},
%   cell{4}{1} = {r=2}{},
%   hline{1,6} = {-}{0.08em},
% }
%  0& 1 & 1 &1  &  1\\
%  6&2  &2  & 2 & 2 \\
%  & 3 & 3 & 3 & 3 \\
%  8&4  &  4& 4 & 4 \\
%  & 5 &  5& 5 &  5
% \end{tblr}
% \end{table}

\begin{table}[t]
\centering
% \small
\footnotesize
\setlength{\tabcolsep}{3.0pt} 
\caption{\textbf{Success rates ($\%$) comparison of different methods.} $\pi_0$ w/ F: $\pi_0$ with native force input. 
% $\pi_0$ w/ I: $\pi_0$ with low-stiffness impedance controller.
Each value represents the success rate over 20 trials. \texttt{Avg.}: the mean performance across all five contact-rich tasks.
\colorbox{lightcyan}{cyan}: main metric.
\textbf{bold}: best results.
}
\vspace{-2mm}

\begin{tabular}{llcccccc}

\toprule
Type&Method & \begin{tabular}[c]{@{}c@{}}Press bottle\end{tabular} & \begin{tabular}[c]{@{}c@{}}Clean vase\end{tabular} & \begin{tabular}[c]{@{}c@{}}Clean board\end{tabular} & \begin{tabular}[c]{@{}c@{}}Retri. plate\end{tabular} & \begin{tabular}[c]{@{}c@{}}Assem. gears\end{tabular} & Avg. \\
\midrule
\multirow{2}{*}{w/o Force}&$\pi_0$
% ~\cite{black2410pi0}
& 35.0 & 20.0 & 35.0 & 0.0 & 0.0 & 18.0 \\
&$\pi_{0.5}$
% ~\cite{pi0.5}
& 45.0 & 30.0 & 45.0 & 15.0 & 20.0 & 31.0 \\
% &$\pi_0$ w/ I& 45.0& 30.0&  35.0& 0.0 & 0.0&22.0 \\

\midrule
\multirow{4}{*}{w/  Force}&ACP
% ~\cite{adaptiveCP}
& 25.0 &30.0 &25.0 & 0.0 & 0.0&16.0 \\
&$\pi_0$ w/ F& 30.0& 25.0&  20.0& 10.0 & 0.0&17.0 \\
&ForceVLA
% ~\cite{yu2025forcevla}
& 70.0 & 25.0 & 55.0 & 15.0 & 10.0& 35.0\\

\rowcolor[rgb]{0.902,1,1} &\textbf{ForceVLA2} & \textbf{80.0} & \textbf{75.0} & \textbf{70.0} & \textbf{35.0}& \textbf{70.0}{\color{teal}$^{\uparrow 50}$}& \textbf{66.0}

\\

\bottomrule
\end{tabular}
\label{main results}
\vspace{-2mm}

\end{table}

\begin{table}[t]
\centering
\small
\setlength{\tabcolsep}{3pt} 
\caption{\textbf{Ablation Study of Different Modules in the ForceVLA2 Model.}
\texttt{FP}: Force Prompt. \texttt{ME}: Multimodal Encoder.  \texttt{CM}: Cross-Scale MoE module.
Entries indicate success rate ($\%$).\colorbox{gray!15}{gray}: baseline results.}
\vspace{-2mm}
\begin{tabular}{ccccccccl}
\toprule
FP & ME & CM & \begin{tabular}[c]{@{}c@{}}Press bottle\end{tabular} & \begin{tabular}[c]{@{}c@{}}Clean vase\end{tabular} & \begin{tabular}[c]{@{}c@{}}Clean board\end{tabular} & \begin{tabular}[c]{@{}c@{}}Retri. plate\end{tabular} & \begin{tabular}[c]{@{}c@{}}Assem. gears\end{tabular} & Avg. \\
\midrule
\grayrow & & & 35.0 & 20.0 & 35.0 & 0.0 &0.0  &18.0  \\
 % &\checkmark&\checkmark&70.0 & 80.0 & 60.0 & 20.0 &  &  \\
 \checkmark &&&60.0 & 25.0 &40.0  &5.0  & 5.0 &27.0{\color{teal}$^{\uparrow 9}$} 
 \\
\checkmark&\checkmark  & &60.0  & 40.0 &65.0  & 5.0 &  30.0&  40.0{\color{teal}$^{\uparrow 13}$}\\

\checkmark & \checkmark& \checkmark & \textbf{80.0} & \textbf{75.0} & \textbf{70.0} & \textbf{35.0} & \textbf{70.0} &\textbf{66.0}{\color{teal}$^{\uparrow 26}$}\\
\bottomrule
\end{tabular}
\label{ablation 1}
% \vspace{-2mm}
\end{table}

\begin{table}[t]
\centering
\small
\setlength{\tabcolsep}{4pt} 
\caption{\textbf{Ablation of input modality combinations in the Cross-Scale MoE module.} 
\texttt{VM}: Visual Modality. 
\texttt{FM}: Force Modality. Entries indicate success rate ($\%$).
% \colorbox{gray!15}{gray}: baseline results. 
% \textbf{bold}: best results.
}
\vspace{-2mm}
\begin{tabular}{ccccccccc}
\toprule
VM & FM   & \begin{tabular}[c]{@{}c@{}}Press bottle\end{tabular} & \begin{tabular}[c]{@{}c@{}}Clean vase\end{tabular} & \begin{tabular}[c]{@{}c@{}}Clean board\end{tabular} & \begin{tabular}[c]{@{}c@{}}Retri. plate\end{tabular} & \begin{tabular}[c]{@{}c@{}}Assem. gears\end{tabular} & Avg. \\
\midrule
 \grayrow & &70.0&30.0  &60.0  &0.0 &20.0  &36.0$^{\quad~}$  \\
\checkmark  & &\textbf{85.0}&40.0  &\textbf{85.0}  &0.0  &40.0  &50.0{\color{teal}$^{\uparrow 14}$}  \\
\checkmark &\checkmark    &80.0  &\textbf{75.0}  &70.0 &\textbf{35.0}  &\textbf{70.0} &\textbf{66.0}{\color{teal}$^{\uparrow 16}$} \\
\bottomrule
\end{tabular}
\label{ablation2}
\vspace{-4mm}
\end{table}

\noindent\textbf{Dynamic force-tracking under sudden perturbations.}
In addition to outperforming baselines on standard benchmarks, ForceVLA2 exhibits exceptional force-tracking performance across a variety of dynamic tests. As depicted in~\cref{Fig: additional exp}, during the bottle pressing task, when the robot arm is about to press the bottle, we abruptly lower the base. Thanks to its force-aware interaction, ForceVLA2 reacts quickly to the new contact configuration and completes the pressing. In contrast, other VLAs slowly chase the new EE 6D pose, leading to failure to maintain stable contact.

\noindent\textbf{Robust wiping and re-targeting in object search.}
In the vase wiping task, when the vase is rotated such that its surface normal becomes misaligned with the arm’s original motion direction, ForceVLA2 drives the arm to slide along the vase and move upward, maintaining contact and successfully completing the wiping motion. Other VLAs, however, continue along their original trajectory, which often results in arm overload and task failure. In the Retrieve Plate task, ForceVLA2 explores the sandbox through force-guided probing with the gripper; even after an initial failed grasp, it autonomously retries, showcasing robust adaptation.

\subsection{Ablation Study}
% In this section, we gradually introduce the modules designed on top of $\pi_0$, including force prompts,Cross-Scale
% MoE module, and the Multimodal
% Encoder. We compare the impact of these different modules across multiple tasks.

% In this section, we progressively introduce the core modules of ForceVLA2, including the force prompts, the Cross-Scale MoE module, and the Multimodal Encoder, and compare their impact across multiple tasks. In particular, we conduct an ablation on the Cross-Scale MoE module by varying its modality inputs and outputs to verify the effectiveness of our design. 

\noindent
% \textbf{Claim 1}.
% 
% As show in \cref{ablation 1}, 
% 

\noindent\textbf{Ablation setup.}
In this section, we progressively introduce the additional modules built on top of $\pi_0$, including the force prompts, the Cross-Scale MoE module, and the Multimodal Encoder, and compare their impact across multiple tasks. In particular, we conduct an ablation on the Cross-Scale MoE module by varying its modality inputs and outputs to verify the effectiveness of our design. Additional experiments focusing on the Multimodal Encoder are provided in the Appendix ~{\color{red} C}.

\noindent\textbf{Component-wise ablations on FP, CM, and ME (Q2).}
As shown in~\cref{ablation 1}, we conduct a stepwise ablation in which we progressively add the Force Prompt (FP), Cross-Scale MoE (CM), and Multimodal Encoder (ME) modules on top of $\pi_0$, and measure the success rate over five tasks. The results show a consistent increase in success rate as more modules are introduced, confirming the effectiveness of our overall architectural design. We can see that the Cross-Scale MoE module provides the most significant gain in success rate among all added components.

\noindent\textbf{Modality fusion in the Cross-Scale MoE module (Q3).}
We further perform an ablation study on the Cross-Scale MoE module to validate our design choices. By selectively enabling or disabling force modality (FM) and visual modality (VM), we examine whether the proposed cross-scale fusion scheme is indeed necessary and beneficial. As reported in~\cref{ablation2}, VM and FM inputs contribute roughly equally within the Cross-Scale MoE module, indicating that both visual–language context and force signals are essential for achieving strong performance. Interestingly, for the Press Bottle and Clean Board tasks, introducing an extra force token on the input side slightly degrades performance. We hypothesize that this effect arises because the force requirements in these tasks are relatively simple, requiring only a single switch between position and force control, so additional force tokens introduce unnecessary degrees of freedom that can perturb an otherwise stable policy, leading to a minor drop in success rate.

% \vspace{-1mm}

\section{Conclusion}
% \vspace{-2mm}

We present ForceVLA2, a force-aware vision-language-action framework with hybrid force–position control for contact-rich manipulation, and evaluate it on our ForceVLA2-Dataset of five real-world tasks.
% : pressing a bottle, cleaning a vase, cleaning a board, retrieving a plate, and assembling gears. 
Across these tasks, ForceVLA2 achieves a success rate of 66\%, outperforming $\pi_0$, $\pi_{0.5}$, and ForceVLA, which obtain 48\%, 35\%, and 31\% on average, and surpassing the second-best method by 50 percentage points on the most force-sensitive assembling gears task. The model also consistently reduces failures caused by arm overload and unstable contact, and ablation studies show monotonically improved performance as force prompts, the Cross-Scale MoE, and multimodal fusion are added, with the full ForceVLA2 variant achieving the best results. 
Overall, ForceVLA2 coupling force–position control achieves more reliable and adaptable contact-rich manipulation than existing state-of-the-art VLA baselines.
% Overall, these experiments demonstrate that explicitly coupling force-aware task concepts with hybrid force–position control yields substantially more reliable and adaptable contact-rich manipulation than existing VLA baselines.

\newpage
\section*{Appendix}
% The appendix provides additional analyses, implementation details, and supplementary results that complement the main paper. Specifically, we include:
% \begin{enumerate}
%     \item Supplementary \textbf{videos} include (1) an overview of the Foca-VLA framework and its core design, and (2) a demonstration of its performance across multiple real-world tasks (\texttt{foca-vla\_project\_video.mp4} and \texttt{foca-vla\_experiment\_video.mp4}).
%     \item Control-Theoretic Analysis of the Reactive Force Pathway (Appendix {\color{red}A}).
%     \item Statistical details of the Foca-Dataset, including distribution characteristics and variance across all tasks (Appendix {\color{red}B}).
%     \item Additional details on training, deployment, and experiments (Appendix {\color{red}C}).
%     \item Discussion on hardware setup and evaluation (Appendix {\color{red}D}).
% \end{enumerate}

\appendix

\section{Theoretical Analysis of Force Pathway}
\label{supp:Control-theoretic analysis}

In this section, we provide a control-theoretic interpretation of why the proposed \textit{reactive force pathway}, the short-horizon reactive manipulation skill that bypasses high-level fusion and feeds force observations directly into the action expert, enhances stability and responsiveness in contact-rich manipulation. As discussed in the main paper, ForceVLA2 routes slowly varying proprioceptive states together with force-aware task knowledge from the VLM expert into the multimodal encoder of the action expert, while injecting rapidly changing force observation directly into the Cross-Scale MoE. This short, high-fidelity gradient path improves the observability and controllability of contact forces under hybrid force–position control, thereby enabling rapid force feedback, reducing reliance on past trajectories, and supporting active, force-guided exploration and interaction.

\noindent
\textbf{\textit{Problem definition.}}
To formally justify this architectural design, we first analyze the fundamental control-theoretic limitations of traditional position-only control. Consider a robotic manipulator interacting with an unknown environment. The proprioceptive state refers to the robot’s end-effector (EE) 6D pose $\mathbf{p} \in \mathbb{R}^7$ (position and quaternion) and interaction force $\mathbf{f} \in \mathbb{R}^6$ (force and torque), it is worth noting that in the control scenario, we utilize EE 6D pose $\mathbf{p_{e}} \in \mathbb{R}^6$ (position and rotation). The critical challenge lies in the environmental dynamics, characterized by an unknown nonlinear mapping:
\begin{equation}
    \mathbf{f} = \boldsymbol{\Phi}(\mathbf{p_{e}}, \dot{\mathbf{p}}_{e}, \boldsymbol{\theta}_e),
\end{equation}
where $\boldsymbol{\Phi}: \mathbb{R}^{12} \times \Theta_e \rightarrow \mathbb{R}^6$ represents the environment's constitutive law with unknown parameters $\boldsymbol{\theta}_e$.

\noindent
\textbf{\textit{Analysis.}}
When using position-only control with force input, the control policy outputs target EE 6D pose  $\mathbf{a_p} \in \mathbb{R}^6$. The system dynamics under ideal position tracking become:
\begin{equation}
\begin{aligned} 
\mathbf{p}(k+1) &= \mathbf{a_p}(k), \\ 
\mathbf{f}(k+1) &= \boldsymbol{\Phi}(\mathbf{a_p}(k), \dot{\mathbf{a}}_p(k), \boldsymbol{\theta}_e). 
\end{aligned}
\end{equation}

The observability matrix for this system has full rank since both $\mathbf{p}_{e}$ and $\mathbf{f}$ are directly measured through vision and force sensors. However, the controllability analysis reveals a fundamental limitation. The controllability matrix for the augmented state $\mathbf{z} = [\mathbf{p}_{e}^T, \mathbf{f}^T]^T \in \mathbb{R}^{12}$ is:
\begin{equation}
\mathcal{C} = [\mathbf{B} \quad \mathbf{A}\mathbf{B} \quad \cdots \quad \mathbf{A}^{11}\mathbf{B}],
\end{equation}
where $\mathbf{A}$ is the system dynamics matrix and the input matrix $\mathbf{B} \in \mathbb{R}^{12\times 6}$ has the structure:
\begin{equation}
\mathbf{B} = \begin{bmatrix} \mathbf{I}_6 \\ \mathbf{0}_{6 \times 6} \end{bmatrix},
\end{equation}
immediately implying:
\begin{equation}
\text{rank}(\mathcal{C}) \leq \text{rank}(\mathbf{B}) = 6 < 12 = \text{dim}(\mathbf{z}).
\end{equation}

\noindent
\textbf{\textit{Conclusion.}}
This rank deficiency arises because force $\mathbf{f}$ is not an independent control variable but rather a dependent output determined by the unknown environment dynamics $\boldsymbol{\Phi}$. Consequently, the reachable set in the task space is constrained to:
\begin{equation}
\mathcal{R} = \{(\mathbf{p}_{e}, \mathbf{f}) : \mathbf{f} = \boldsymbol{\Phi}(\mathbf{p}_{e}, \dot{\mathbf{p}}_{e}, \boldsymbol{\theta}_e), \quad \mathbf{p}_{e} \in \mathcal{P}\},
\end{equation}
which represents a 6-dimensional manifold embedded in the 12-dimensional task space, confirming that arbitrary force-position combinations cannot be achieved through position-only control.

\noindent
\textbf{\textit{Innovation.}}
ForceVLA2 fundamentally transforms the control architecture by introducing hybrid force-position control $\mathbf{a} = [\mathbf{a}_p^T, \mathbf{f}_{\text{t}}^T]^T \in \mathbb{R}^{12}$. The key innovation lies in employing a deep neural network policy $\pi_\theta: \mathcal{O} \rightarrow \mathbb{R}^{12}$ that learns from interaction observations.

Through supervised learning on expert demonstrations, the policy implicitly acquires an inverse model of the environment dynamics. Specifically, for a desired task-space target $(\mathbf{p}_e^*, \mathbf{f}^*)$, the trained policy learns to compute control commands that achieve this target by leveraging the learned relationship between positions and forces. The policy effectively learns to solve the inverse problem, i.e., $\text{given } \mathbf{f}^*=\boldsymbol{\Phi}(\mathbf{p}_{e}^*, \dot{\mathbf{p}}_{e}^*, \boldsymbol{\theta}_e), $
it finds control inputs that achieve $(\mathbf{p}_e^*, \mathbf{f}^*).$

The learned policy effectively extends the control authority to approach the full 12-dimensional action space. By incorporating both position and force control, and learning their coupling through the environment dynamics, Foca-VLA can achieve force-position combinations that lie beyond the restricted manifold of position-only control. The key insight is that the policy learns to coordinate position and force commands in a way that respects the physical constraints while expanding the reachable set.

To quantify this enhancement, we define the effective controllability index:
\begin{equation}
\kappa = \frac{\text{dim}(\mathcal{R})}{\text{dim}(\mathcal{Z})},
\end{equation}
where $\mathcal{R}$ is the reachable set and $\mathcal{Z}$ is the task space. For position-only control, $\kappa_{\text{position-only}} = 6/12 = 0.5$, indicating that only 50\% of the task space dimensionality is controllable. With Foca-VLA and a well-trained policy, the reachable set expands significantly as the policy learns to exploit the learned environment model, leading to $\kappa_{\text{FocaVLA}} \to \kappa_{\text{max}}$, where $\kappa_{\text{max}}$ depends on the richness of the training data and the physical constraints of the task.

This improvement from $\kappa =0.5$ toward higher values represents the fundamental achievement of Foca-VLA: leveraging learned environment knowledge to expand the controllable subspace and enable precise force-position coordination in contact-rich manipulation tasks. The reactive force pathway further enhances this capability by providing rapid, direct force feedback to the action expert, enabling active adjustment of interaction forces during contact-rich manipulation.

\section{Statistical details of the Foca-Dataset}
\label{supp:Statistical details of the Foca-Dataset}
\begin{figure*}[t]
\centering
\includegraphics[width=\textwidth]{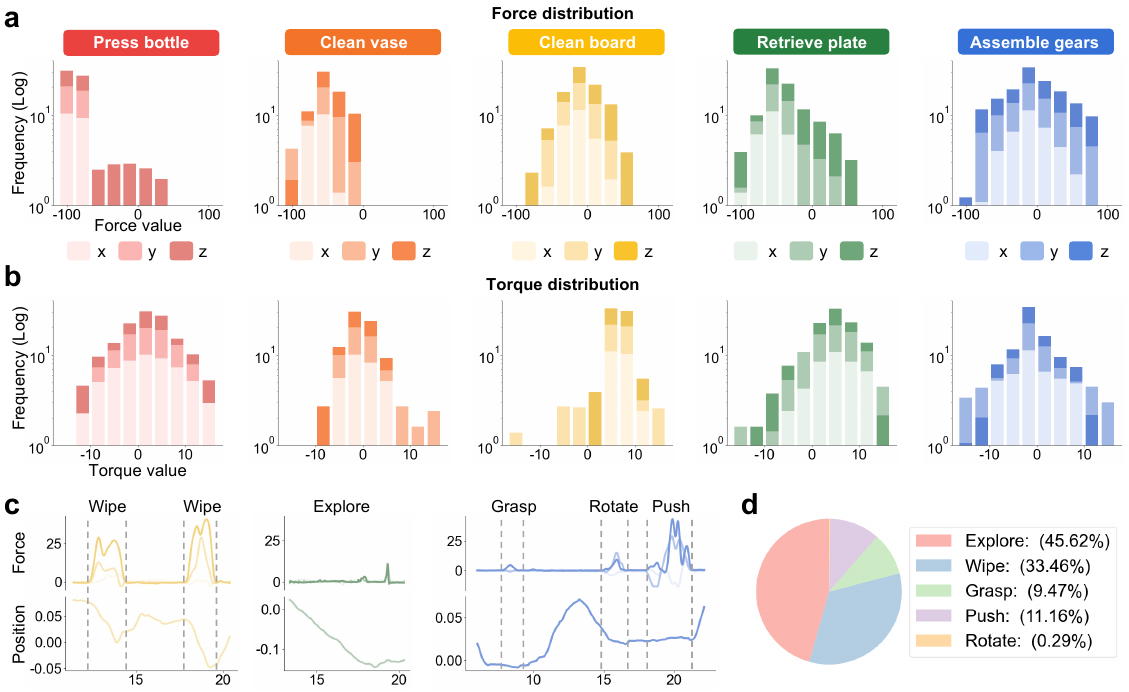}
\caption{
\textbf{Statistical details of ForceVLA2-Dataset.} Frequency distributions of force components (a) and (b) torque components of $x$, $y$, $z$ dimensions recorded during task execution. (c) Illustration of short-horizon reactive manipulation skills based on force and position. (d) Skill distribution.
}
\label{Fig: Dataset}
\end{figure*}

We analyze the statistical distribution of normalized forces (scaled to $\pm$100 N) and normalized torques (scaled to $\pm$15 Nm) across manipulation tasks in the Foca-Dataset, which exhibit task-specific characteristics (\cref{Fig: Dataset} (a) and (b)). 

For the force distribution, forces in the \texttt{Press bottle} task are predominantly concentrated along the negative $z$-axis (vertical downward direction). 
The \texttt{Clean vase} and \texttt{Clean board} tasks show more distributed force patterns across all three axes, reflecting the wiping motions that involve both tangential sliding forces and normal contact forces. The \texttt{Retrieve plate} task demonstrates relatively symmetric distributions centered near zero across all axes, suggesting exploratory probing behaviors with gentle contact forces. The \texttt{Assemble gears} task exhibits a broader force distribution, particularly along the $z$-axis, corresponding to actions during the alignment and insertion phases.

For the torque distribution, the \texttt{Press bottle} task shows concentrated torque distributions near zero across all axes, indicating minimal rotational adjustments. In contrast, the \texttt{Clean vase}, \texttt{Clean board}, and \texttt{Retrieve plate} tasks exhibit wider torque spreads, reflecting the continuous orientation adjustments required for grasping, wiping, and probing. Moreover, the \texttt{Assemble gears} task presents the most diverse torque distribution, particularly in the $z$-axis, corresponding to the precise rotational alignment required during gear insertion.

Furthermore, we decompose the five tasks into short-horizon reactive manipulation skills. As illustrated in \cref{Fig: Dataset} (c), we identify five fundamental skills based on force and position trajectories. For the Wipe skill, frames during the period where position trajectories change more than 0.05 m and force amplitude is more than 10 N. For the Push skill, frames during the period where $z$-axis position trajectories change more than 0.05 m and $z$-axis force amplitude more than 5 N. For the Grasp skill, frames during the period where $z$-axis position trajectories change more than 0.1 m and force amplitude more than 5 N. For the Rotate skill, frames during the period where forces of three axes change more than 1 N. Frames outside the above definitions are the Explore skills, featured with low-magnitude force and rapid position drift. Statistical analysis shows that the Foca-Dataset is dominated by Explore (45.62\%, followed by Wipe, Push, Grasp and Rotate, indicating diverse force-position coordination patterns.

\section{Implementation and Experiments}
\label{sec:Training details and More Experiment}

\begin{figure*}[t]
\centering
\includegraphics[width=\textwidth]{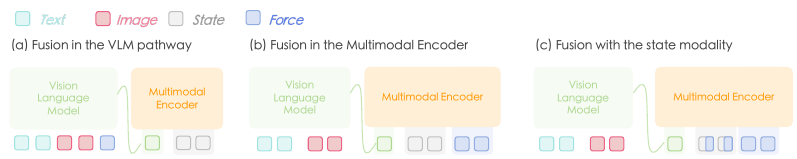}
\caption{
% \TR{
\textbf{Ablation study of force injection in the multimodal encoder.} The figure illustrates three candidate insertion points for the force branch: (1) fusion in the VLM pathway, (2) fusion in the multimodal encoder, and (3) fusion with the state, building upon (2).
}
\label{Fig: ablation 4}
\vspace{-5mm}
\end{figure*}

In this section, we provide a detailed explanation of our training strategy and hyperparameters. Additionally, we include an ablation study on the multimodal encoder.
As shown in \cref{table: training details}, we trained the Foca-VLA model on 8 A100 GPUs with a batch size of 32 and 30{,}000 total training steps, which took approximately 10 hours. Foca-VLA achieves an inference speed of 15 Hz on a 4090 GPU with a chunk size of 30.
\begin{table}[]

\centering
\small
\setlength{\tabcolsep}{10pt} 
\caption{\textbf{Training details of Foca-VLA.}
}
\centering
\begin{tabular}{lc}

\toprule
\textbf{Parameter} & \textbf{Value} \\
\midrule
Learning Rate Schedule & Cosine Decay \\
Optimizer & AdamW \\
EMA Decay & 0.99 \\

Random Seed & 42 \\
Batch Size & 32 \\

Training Steps & 30,000 \\

\bottomrule
\end{tabular}
\label{table: training details}
\end{table}

% 需要在导言区引入：
% \usepackage{booktabs}

In the multimodal encoder module, we further investigate where to inject the force branch to validate its effectiveness. 
As illustrated in \cref{Fig: ablation 4}, there are three candidate insertion points: (1) into the VLM pathway, (2) into the multimodal encoder, and (3) fused jointly with the state ( EE 6D pose). The third insertion point (3) is the fusion of force with the state, building upon the second point (injection into the multimodal encoder) and combining it with the state model. As reported in \cref{ablation 4}, injecting force into the VLM pathway leads to a substantial degradation in overall performance, while injecting it into either the multimodal encoder or the state fusion branch yields consistent gains. Based on these observations, our final Foca-VLA design incorporates force at both the multimodal encoder and state fusion levels.

\begin{table}[]
\label{table: ablation study on multimodal encoder}
\centering
\small
\setlength{\tabcolsep}{3.8pt} 
\caption{\textbf{Success rates ($\%$) of kinds of force injection in the multimodal encoder.}
\texttt{ME}: Multimodal Encoder. State fusion means fusion with state modality, building upon ME.
}
\vspace{-2mm}
\begin{tabular}{lcccccc}
\toprule
Method & \begin{tabular}[c]{@{}c@{}}Press bottle\end{tabular} & \begin{tabular}[c]{@{}c@{}}Clean vase\end{tabular} & \begin{tabular}[c]{@{}c@{}}Clean board\end{tabular} & \begin{tabular}[c]{@{}c@{}}Retri. plate\end{tabular} & \begin{tabular}[c]{@{}c@{}}Assem. gears\end{tabular} & Avg. \\
\midrule
VLM Pathway & 10.0 & 10.0  &   5.0& 0.0  &   0.0& 5.0  \\
ME &75.0 & 55.0  & 65.0 & 40.0 & 55.0  & 58.0 \\
\midrule
\rowcolor[rgb]{0.902,1,1} \textbf{State Fusion} & \textbf{80.0} & \textbf{75.0} & \textbf{70.0} & \textbf{35.0}& \textbf{70.0}& \textbf{66.0}\\
\bottomrule
\end{tabular}
\label{ablation 4}
\vspace{-2mm}
\end{table}

Results show that injecting force into the VLM pathway causes significant performance degradation, while force integration into the Multimodality Encoder or EE 6D pose fusion branch consistently improves model performance.

\section{Discussion on hardware and evaluation.}
\label{sec: Discussion on hardware and evaluation}

\textbf{Hardware-agnostic setup.} 
    We adopt a standard, hardware-agnostic setup. By grounding execution in a Jacobian-based mapping, Foca-VLA decouples the policy from kinematics, enabling compatibility with: (i) torque-controlled robots (\textit{e.g.}, Franka); (ii) robots with EE F/T sensors (\textit{e.g.}, UR); and (iii) torque-interface actuators (\textit{e.g.}, Feetech), allowing low-cost deployment.
% \noindent

% \textbf{\textcolor{myblue}{[EXPL3]}} 
\textbf{Real-world evaluation.} We focus on real-world experiments because force interactions are sensitive to friction/contact modeling, which makes simulation results less reliable for our setting. In addition, there is no widely available benchmark that is directly usable for force-aware VLAs; therefore, real-robot evaluations provide the most direct evidence.

\clearpage
{
    \small
    \bibliographystyle{tips/ieeenat_fullname}
    \bibliography{main}
    
}
% \input{sec/supp}


\end{document}